\documentclass{article}

\usepackage[dvipsnames]{xcolor}
\usepackage{tabularx}
\usepackage{tikz}
\usetikzlibrary{fit,calc}
\newcommand*{\tikzmk}[1]{\tikz[remember picture,overlay,] \node (#1) {};\ignorespaces}
\newcommand{\boxit}[1]{\tikz[remember picture,overlay]{\node[yshift=3pt,fill=#1,opacity=.25,fit={(A)($(B)-(0, 0.5\baselineskip)$)}] {};}\ignorespaces}

\newcommand{\rchi}{\mathcal{X}}
\newcommand{\data}{\mathcal{D}}
\newcommand{\batch}{\hat{x}}
\newcommand{\keloss}{\tilde{\ell}}

\newcommand{\client}{k}
\newcommand{\clients}{K}
\newcommand{\kl}{\mathsf{KL}}
\newcommand{\ce}{\mathsf{CE}}
\usepackage{amsmath}
\DeclareMathOperator*{\argmax}{arg\,max}
\DeclareMathOperator*{\argmin}{arg\,min}

\newcommand{\expectation}{\mathop{\mathbb{E}}}

\usepackage{xspace}
\newcommand{\codream}{\texttt{CoDream}\xspace}
\newcommand{\feddream}{\texttt{CoDream}\xspace}
\newcommand{\fastfeddream}{\texttt{CoDream-fast}\xspace}
\newcommand{\fedavg}{\texttt{FedAvg}\xspace}

\newcommand{\independent}{\texttt{Independent}\xspace}
\newcommand{\centralized}{\texttt{Centralized}\xspace}

\usepackage{microtype}
\usepackage{graphicx}
\usepackage{subfigure}
\usepackage{booktabs} 
\usepackage{soul}
\usepackage{algorithm2e}
\usepackage{hyperref}

\usepackage{color, colortbl}

\usepackage[accepted]{icml2024}

\usepackage{amsmath}
\usepackage{amssymb}
\usepackage{mathtools}
\usepackage{amsthm}
\usepackage[capitalize,noabbrev]{cleveref}

\theoremstyle{plain}

\theoremstyle{definition}

\theoremstyle{remark}

\usepackage[textsize=tiny]{todonotes}

\icmltitlerunning{CoDream: Exchanging dreams instead of models for federated aggregation with heterogeneous models}

\begin{document}

\twocolumn[
\icmltitle{CoDream: Exchanging dreams instead of models for federated aggregation with heterogeneous models}



\icmlsetsymbol{equal}{*}

\begin{icmlauthorlist}
\icmlauthor{Abhishek Singh}{equal,mit}
\icmlauthor{Gauri Gupta}{equal,mit}
\icmlauthor{Ritvik Kapila}{ucsd}
\icmlauthor{Yichuan Shi}{mit}
\icmlauthor{Alex Dang}{mit}
\icmlauthor{Sheshank Shankar}{mit}
\icmlauthor{Mohammed Ehab}{mit}
\icmlauthor{Ramesh Raskar}{mit}
\end{icmlauthorlist}

\icmlaffiliation{mit}{Massachusetts Institute of Technology, USA}
\icmlaffiliation{ucsd}{University of California San Diego, USA}

\icmlcorrespondingauthor{Abhishek Singh}{abhi24@mit.edu}
\icmlcorrespondingauthor{Gauri Gupta}{gaurii@mit.edu}

\icmlkeywords{Machine Learning, ICML}

\vskip 0.3in
]



\printAffiliationsAndNotice{\icmlEqualContribution}

\begin{abstract}
Federated Learning (FL) enables collaborative optimization of machine learning models across decentralized data by aggregating model parameters. Our approach extends this concept by aggregating ``knowledge" derived from models, instead of model parameters. We present a novel framework called \codream, where clients collaboratively optimize randomly initialized data using federated optimization in the input data space, similar to how randomly initialized model parameters are optimized in FL. Our key insight is that jointly optimizing this data can effectively capture the properties of the global data distribution. Sharing knowledge in data space offers numerous benefits: (1) model-agnostic collaborative learning, i.e., different clients can have different model architectures; (2) communication that is independent of the model size, eliminating scalability concerns with model parameters; (3) compatibility with secure aggregation, thus preserving the privacy benefits of federated learning; (4) allowing of adaptive optimization of knowledge shared for personalized learning. We empirically validate \codream on standard FL tasks, demonstrating competitive performance despite not sharing model parameters. Our code: \url{https://mitmedialab.github.io/codream.github.io/}
\end{abstract}
\section{Introduction}
\begin{figure*}[h]
    \centering
    \includegraphics[width=\textwidth]{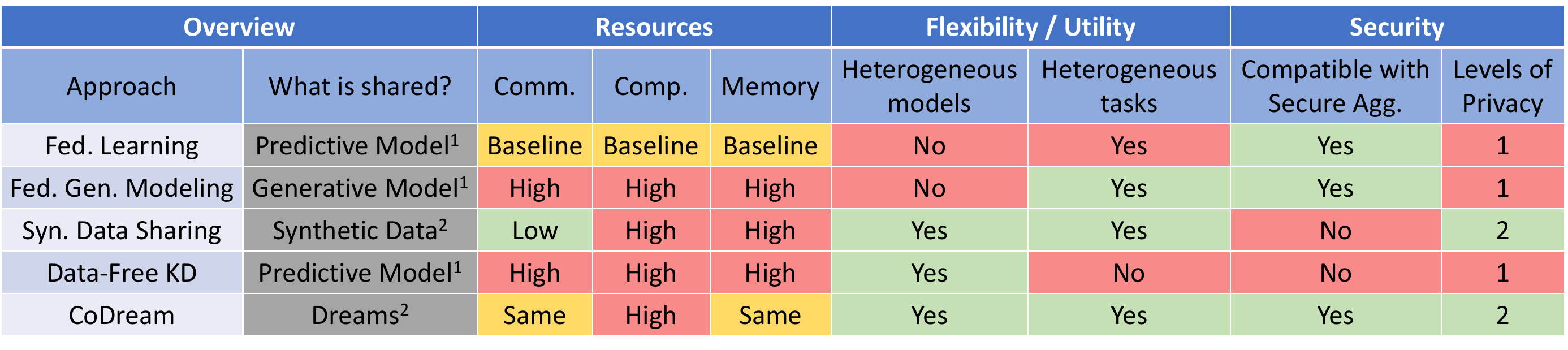}
     \vspace{-25pt}
    \caption[]{{\textbf{Landscape of FL techniques.} Here we use Fed.-Federated, Gen.-Generative, Syn.-Synthetic, Pred.-Predictive, Comm.-Communication, Comp.-Computation, Het.-Heterogeneous, Agg.-Aggregation. By levels of privacy, we mean how distant the shared updates are from raw data. Sharing synthetic data\footnotemark[2] and \textit{dreams}\footnotemark[2] are two levels of indirection away from the raw data than sharing models\footnotemark[1]. }}
    \vspace{-10pt}
    \label{fig:overviewFL}
\end{figure*}
In many application areas, such as healthcare and finance, data is distributed among silos owned by different organizations, making it difficult to train machine learning (ML) models on large datasets collaboratively. Centralizing data is not always feasible due to regulatory and privacy concerns. Federated Learning (FL)~\cite{mcmahan2023communicationefficient} addresses this problem by centrally aggregating clients' models instead of their data. Informally, FL circumvents privacy concerns in two steps: 1) sharing client's models instead of data offers confidentiality as data does not leave the trusted local device, and 2) the aggregation step in FL is a linear operation (weighted average), which makes it compatible with secure aggregation techniques. The efficacy of the first layer of privacy becomes pronounced when the number of samples per client is substantial, while the significance of the second layer becomes apparent in ecosystems with numerous clients.

FL assumes that all clients agree on the same model architecture and are willing to share their local models. Due to resource constraints, however, this can potentially reduce the number of clients in the ecosystem, eliminating the benefit of the second layer. Some recent knowledge-distillation (KD)~\cite{mora2022knowledge} techniques present an alternate paradigm that allows clients to share knowledge while allowing heterogeneous models. However, these KD algorithms depart from the model averaging paradigm, making them incompatible with secure aggregation. Hence, they also can not derive privacy benefits from the second layer.

Alternatively, if we could generate samples representing data distribution characteristics while maintaining privacy benefits at both layers, we would eliminate the need to aggregate the client models. Sharing samples offers much higher flexibility for training models and supports arbitrary model architectures and tasks. However, the problem is challenging because collaboratively learning a generative model leads to the same problems as FL for predictive models.
Instead of learning a generative model, we solve this dilemma by optimizing data collaboratively instead of parameters.

We design a novel framework for collaboratively synthesizing a proxy of siloed data distributions without centralizing data or client models. These collaboratively synthesized representations of data, which we call \textit{dreams}, can be used to train ML models. We show that dreams capture the knowledge embedded within local models and also facilitate the aggregation of local knowledge without ever sharing the raw data or models. Our key idea is to begin with randomly initialized samples and apply federated optimization on these samples to extract knowledge from the client's local models trained on their original dataset. Unlike synthetic data, the goal of optimizing \textit{dreams} is to enable KD, rather than generate realistic data (maximize likelihood of data).


We design our framework into three stages: knowledge extraction~\ref{sec:method:k-ex}, knowledge aggregation~\ref{sec:method:k-ag}, and knowledge acquisition~\ref{sec:method:k-ac}. We perform extensive investigation to test \feddream by (1) establishing the feasibility of \feddream as a way for clients to synthesize samples collaboratively, (2) showing the utility of synthesized samples by learning predictive models (3) validating \feddream as an alternative to FL and (4) performing empirical validation of our framework by benchmarking with existing algorithms and ablation studies across various design choices.


The key factors of our approach are: 
(1) \textbf{Flexibility}: Our proposed technique, \feddream, collaboratively optimizes \textit{dreams} to aggregate knowledge from the client's local models. By sharing \textit{dreams} in the data space rather than model parameters, our method is model-agnostic. (2) \textbf{Scalability}: Furthermore, communication does not depend on the model parameter size, alleviating scalability concerns. (3) \textbf{Privacy}: Just like \fedavg~\cite{FL}, \feddream also exhibits two-fold privacy: Firstly, clients share \textit{dreams'} updates instead of raw data. Secondly, the linearity of the aggregation algorithm allows clients to securely aggregate their \textit{dreams} without revealing their individual updates to the server.
In summary, our contributions are as follows:
\begin{itemize}
    \vspace{-10pt}
    \item A novel framework \feddream, for collaborative data synthesis through federated optimization in input space, serving as a proxy for the global data distribution.
    \vspace{-5pt}
    \item Our approach introduces a novel perspective to FL by aggregating ``knowledge" instead of local model parameters. This unique aggregation framework leads to model-agnostic learning and addresses scalability concerns while preserving privacy through compatibility with secure aggregation.
    \vspace{-5pt}
    \item Extensive empirical validation of \codream, including benchmarking against existing algorithms and ablation studies across various design choices, further emphasizes its potential for collaborative optimization and adaptability for personalized learning.

    
    
\end{itemize}

\section{Related Work}
\vspace{-5pt}
The problem of collaborative data synthesis has been previously explored using generative modeling and federated learning techniques. Figure~\ref{fig:overviewFL} compares existing decentralization solutions regarding shared resources, utility, and privacy. We refer the reader to Supplementary for a more detailed discussion of existing works.

\textbf{Generative modeling} techniques either pool locally generated data on the server~\cite{song2022federated, goetz2020federated} or use \fedavg with generative models~\cite{rasouli2020fedgan, 9054559}. Like FL, \fedavg over generative models is also not model agnostic. While we share the idea of generative data modeling, we do not expose individual clients' updates or models directly to the server.

\footnotetext[1]{Aggregation of local updates occurs in model parameter space}
\footnotetext[2]{Aggregation of local updates occurs in the data space}

\textbf{Knowledge Distillation in FL} is an alternative to \fedavg that aims to facilitate knowledge sharing among clients that cannot acquire this knowledge individually~\cite{chang2019cronus, lin2020ensemble, afonin2022towards, chen2021fedbe}. However, applying KD in FL is challenging because the student and teacher models need to access the same data, which is difficult in FL settings. 

\textbf{Data-free Knowledge Distillation} algorithms address this challenge by employing a generative model to generate synthetic samples as substitutes for the original data ~\cite{zhang2022dense,zhang2022fine,zhu2021data}. 
These data-free KD approaches are not amenable to secure aggregation and must use the same architecture for the generative model.

However, all these existing approaches lack active client collaboration in the knowledge synthesis process. Clients share their local models or locally generated data with the server without contributing to knowledge synthesis.  We believe that collaborative synthesis is crucial for secure aggregation and bridging the gap between KD and FL. Our approach \feddream enables clients to synthesize dreams collaboratively while remaining compatible with secure aggregation techniques and being model agnostic.

\section{Preliminaries}
\textbf{Federated Learning (FL)} aims to minimize the expected risk $\min_{\theta}\mathbb{E}_{\data\sim p(\data)} \ell(\data, \theta)$ where $\theta$ is the model parameters, $\data$ is a tuple of samples $\left(X\in\rchi, Y\in\mathcal{Y}\right)$ of labeled data in supervised learning in the data space $\rchi\subset\mathbb{R}^d$ and $\mathcal{Y}\subset\mathbb{R}$, and $\ell$ is some risk function such as mean square error or cross-entropy~\cite{konevcny2016federated, mcmahan2023communicationefficient}. In the absence of access to the true distribution, FL aims to optimize the empirical risk instead given by:
\vspace{-5pt}
\begin{equation}
    \min_{\theta}\sum_{\client\in \clients} \frac{1}{|\data_\client|} \ell(\data_\client, \theta),
    \label{eq:fl}
    \vspace{-5pt}
\end{equation}
$\data$ is assumed to be partitioned across $\clients$ clients, where each client $\client$ owns each $\data_\client$ and  $\data=\cup_{\client\in\clients} \data_\client$. The optimization proceeds with the server broadcasting $\theta^r$ to each user $k$ that locally optimizes $\theta_k^{r+1}=\argmin_{\theta^r}\ell(\data_\client, \theta^r)$ for $M$ rounds and sends local updates either in the form of $\theta_\client^{r+1}$ or $\theta_\client^{r+1} - \theta_\client^r$ (\textit{pseudo-gradient}) to the server to aggregate local updates and send the aggregated weights back to the clients. \\
\textbf{Knowledge Distillation} facilitates the transfer of knowledge from a teacher model ($f(\theta_T)$) to a student model ($f(\theta_S)$) by incorporating an additional regularization term into the student's training objective~\cite{buciluǎ2006model, hinton2015distilling}. This regularization term (usually computed with Kullback-Leibler (KL) divergence $\kl(f(\theta_T, \data)||f(\theta_S, \data))$) encourages the student's output distribution to match the teacher's outputs. \\ 

\vspace{-10pt}
\textbf{DeepDream for Knowledge Extraction} ~\cite{mordvintsev2015inceptionism} first showed that features learned in deep learning models could be \textit{extracted} using gradient-based optimization in the feature space. Randomly initialized features are optimized to identify patterns that maximize a given activation layer. Regularization such as TV-norm and $\ell_1$-norm has been shown to improve the quality of the resulting images. Starting with a randomly initialized input $\batch\sim\mathcal{N}(0, \; I)$, label $y$, and pre-trained model $f_\theta$, the optimization objective is 
\vspace{-5pt}
\begin{equation}
    \min_{\batch} \ce\left(f_\theta(\batch), \; y\right) \; + \; \mathcal{R}(\batch),
    \vspace{-5pt}
    \label{eq:deepdream}
\end{equation}
where $\ce$ is cross-entropy and $\mathcal{R}$ is some regularization. DeepInversion~\cite{yin2020dreaming} showed that the knowledge distillation could be further improved by matching batch normalization statistics with the training data at every layer.

\section{CoDream}
\label{sec:method}
\begin{figure*}[]
    \centering
    \includegraphics[width=0.90\textwidth]{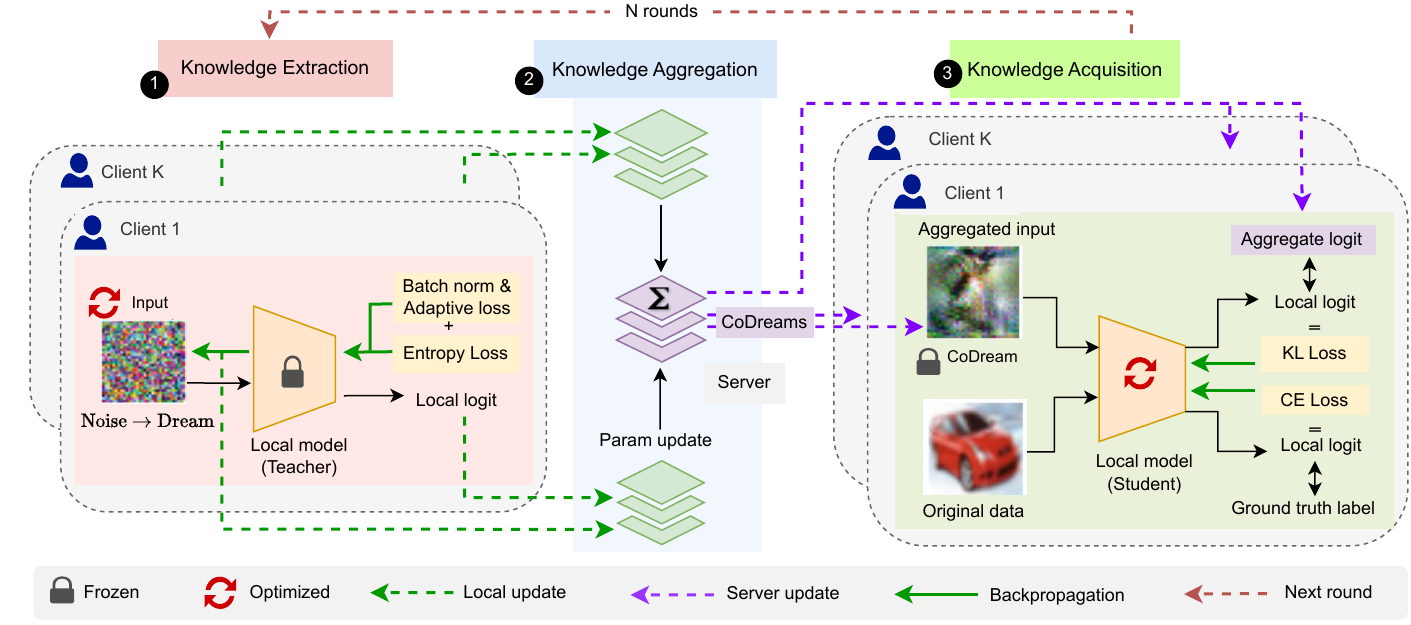}
     \vspace{-10pt}
    \caption[]{\textbf{Overview of the \codream pipeline} comprising three stages: (1) Knowledge Extraction— each client generates \textit{dreams}, representing the extracted knowledge from their local models (teacher). Starting with random noise images and frozen teacher models, clients optimize to reduce entropy on the output distribution while regularizing the batch norm and adaptive loss. The clients share their local updates of \textit{dreams} and logits with the server. (2) Knowledge Aggregation—server aggregates \textit{dreams} and soft labels from clients to construct a \feddream dataset. (3) Knowledge Acquisition—clients update their local models through two-stage training (i) on jointly optimized \textit{co-dreams} with knowledge distillation (where clients act as students) and (ii) local dataset with cross-entropy loss.}
    \vspace{-10pt}
    \label{fig:pipeline}
\end{figure*}

Our approach \feddream consists of three key stages: \textbf{knowledge extraction}, \textbf{knowledge aggregation} and \textbf{knowledge acquisition}. In the knowledge extraction stage, each client extracts useful data representations, referred to as ``dreams'', from their locally trained models (teachers). Starting with random noise images and fixed teacher models, clients optimize these images to facilitate knowledge sharing from their local models (Section \ref{sec:method:k-ex}). Since this is a gradient-based optimization of the input \textit{dreams}, 
we exploit the linearity of gradients to enable knowledge 
aggregation from all the clients. In the knowledge aggregation stage, the clients now jointly optimize these random noised images by aggregating the gradients from the local optimizations (Section \ref{sec:method:k-ag}). Unlike traditional federated averaging (\fedavg), our aggregation occurs in the input data space over these \textit{dreams}, making our approach compatible with heterogeneous client architectures. Finally, in the knowledge acquisition step, these collaboratively optimized images, or \textit{dreams}, are then used for updating the server and clients without ever sharing the raw data or models. This is done by performing knowledge distillation on the global \textit{dreams} where clients now act as students (Section \ref{sec:method:k-ac}). Figure \ref{fig:pipeline} gives an overview of the \feddream pipeline for each round. We further discuss these stages in more detail in the following subsections.

\subsection{Local dreaming for extracting knowledge from models}
\label{sec:method:k-ex}
In this stage, clients perform local \textit{dreaming}, a model-inversion approach to extract useful information from the locally trained models. We use DeepDream~\cite{mordvintsev2015inceptionism} and DeepInversion~\cite{yin2020dreaming} approaches that enable data-free knowledge extraction from the pre-trained models. However, these are not directly applicable to a federated setting because the client models are continuously evolving, as they learn from their own data as well as other clients. A given client should synthesize only those \textit{dreams} over which they are highly confident. As the client models evolve, their confidence in model predictions also changes over time. A direct consequence of this non-stationarity is that it is unclear how the label $y$ should be chosen in Eq~\ref{eq:deepdream}. In DeepInversion, the teacher uniformly samples $y$ from its own label distribution because the teacher has the full dataset. However, in the federated setting, data is distributed across multiple clients with heterogeneous data distributions.

To keep track of a given client's confidence, we take a simple approach of treating the entropy of the output distribution as a proxy for the teachers' confidence. We adjust Eq~\ref{eq:deepdream} so that the teacher synthesizes \textit{dreams} without any classification loss by instead minimizing the entropy (denoted by $\mathcal{H}$) on the output distribution. Each client (teacher) starts with a batch of representations sampled from a standard Gaussian ($\hat x=\mathcal{N}(0, 1)$), and optimizes \textit{dreams} using Eq~\ref{eq:knowledge_extraction}. Formally, we optimize the following objective for synthesizing \textit{dreams}:
\vspace{-5pt}
\begin{equation}
    \label{eq:knowledge_extraction}
    \min_{\batch} \left\{\tilde{\ell}(\batch,\; \theta)\:= \; \mathcal{H}\left(f_{\theta}(\batch)\right) + \mathcal{R}_{bn}(\batch) + \mathcal{R}_{adv}(\batch)\right\}
\end{equation}
where $\mathcal{H}$ is the entropy for the output predictions, $\mathcal{R}_{bn}$ is the feature regularization loss and $\mathcal{R}_{adv}$ is a student-teacher adversarial loss. $\mathcal{R}_{adv}$ helps extract knowledge from the clients that the clients know and the server does not know. More details for these components are provided in the Supplementary section \ref{sec:loss_components}.
Note that generative models create synthetic data with objectives to resemble the real data and align with the input distribution by maximizing the likelihood of the data. Unlike synthetic data, the only goal of optimizing \textit{dreams} is to enable efficient knowledge distillation.  Therefore, \textit{dreams} do not need to appear like real images. We also visualize \textit{dreams} and compare them against real images in Figure \ref{fig:dreams}.


\vspace{-5pt}
\subsection{Collaborative dreaming for knowledge aggregation}
\label{sec:method:k-ag}
Since the data is siloed and lies across multiple clients, we want to extract the collective knowledge from the distributed system. While \fedavg aggregates gradients of the model updates from clients, it assumes the same model architecture across clients and thus is not model-agnostic. 

\begin{figure}[!hbt]
    \centering
    \includegraphics[width=0.5\textwidth]{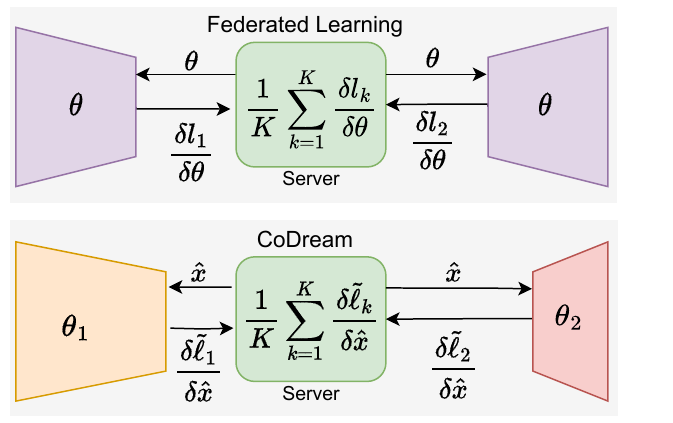}
    \vspace{-20pt}
    \caption{\textbf{Comparing aggregation framework in FL and \feddream.} In FL, the server aggregates the gradients of model parameters, whereas, in \feddream, aggregation happens in the gradients of the data space, called dreams ($\hat{x}$), allowing for different model architectures. Here $K$ is the number of clients and $l, \tilde{l}$ are loss functions given in Eq \ref{eq:fl} and Eq \ref{eq:knowledge_extraction}.}
    \vspace{-6pt}
    \label{fig:FLvsCoDream}
\end{figure}

We propose a novel mechanism for aggregating the knowledge by collaboratively optimizing \textit{dreams} across different clients. Instead of each client independently synthesizing \textit{dreams} using Eq~\ref{eq:knowledge_extraction}, they now collaboratively optimize them by taking the expectation over each client's local loss w.r.t. the same $\batch$: $\min_{\batch}\expectation_{k \in K}\left[\tilde{\ell}(\batch,\; \theta_k) \right]$
This empirical risk can be minimized by computing the local loss at each client. Therefore, the update rule for $\batch$ can be written as:
\vspace{-5pt}
$$\batch \leftarrow \batch - \nabla_{\batch}\sum_{k\in K}\frac{1}{|\data_k|}\tilde{\ell}(\batch,\; \theta_k)$$
Using the linearity of gradients, we can write it as 
\vspace{-5pt}
\begin{equation}
    \batch \leftarrow \batch - \sum_{k\in K}\frac{1}{|\data_k|}\nabla_{\batch}\tilde{\ell}(\batch,\; \theta_k)
    \vspace{-5pt}
    \label{eq:fedream}
\end{equation} 
The clients compute gradients locally with respect to the same input and share them with the server, which aggregates the gradients and returns the updated input to the clients. This formulation is the same as the distributed-SGD formulation, but the optimization is performed in the data space instead of the model parameter space. Thus, unlike \fedavg, our approach \feddream is model-agnostic and allows clients with heterogenous model architecture as shown in Fig \ref{fig:FLvsCoDream}. Our framework is also compatible with existing cryptographic aggregation techniques, as the aggregation step is linear and only reveals the final aggregated output without exposing individual client gradients. 

Collaboratively optimizing representations, known as \textit{dreams} in our approach, is a novel concept that has not been explored before. Our experiments in Section ~\ref{sec:ind_vs_collab} demonstrate that dreams obtained through this approach capture knowledge from all clients and outperform dreams independently synthesized dreams by clients.



\subsection{Knowledge acquisition}
\label{sec:method:k-ac}
Finally, the extracted knowledge, in the form of collaboratively trained \textit{dreams}, is then acquired by the local client and server models to update their models with the global information. They learn to match the distribution of the ensemble of clients on the distilled samples (denoted by $\hat{\data}$), which are obtained by Eq~\ref{eq:fedream}. The clients share soft logits for each \textit{dream}, which are then aggregated by the server to perform knowledge distillation (where clients and server now act as students) on the following training objective:
\vspace{-5pt}
\begin{equation}
    \min_{\theta}\sum_{\batch \in \hat{\data}}\kl\left(\sum_k\frac{1}{|\data_k|}f_{\theta_k}(\batch)\; \bigg\| \; f_{\theta}(\batch)\right)\\
    \label{eq:k-acq}
\end{equation} 
We provide the complete algorithm of \codream in Algorithm \ref{alg:codream}. Note that the choice of parameters such as local updates $M$, global updates $R$, local learning rate $\eta_l$, global rate $\eta_g$, and the number of clients $K$ typically guide the trade-off between communication efficiency and convergence of the optimization.

\begin{algorithm}[h]
    \SetAlgoLined
    \begin{tabularx}{\linewidth}{@{}c @{}X}
          \textbf{Input: } & Number of client K, local models and data $\theta_k \text{ and } \data_k, k \in K$, local learning rate $\eta_k$,  global learning rate $\eta_g$, local training rounds M, global training epochs R, total number of epochs N.  
        \end{tabularx} \\ 
    \For{$t=1$ \KwTo $N$}{
        Server initializes a batch of dreams as $\batch \sim \mathcal{N}(0, 1)$; \\
        \For{$r=1$ \KwTo $R$}{
            Server broadcasts current dream $\hat{x}^r$ to all clients \\
            \For{\text{each client} $k \in K$ \textbf{in parallel}}{
                $\batch^r_{k, 0}:=\batch^{r}$; \\
                \SetInd{0.5em}{0.85em}
                \For{$m=1$ \KwTo $M$}{
                    // Local knowledge extraction stage (Eq \ref{eq:knowledge_extraction})\\
                    \tikzmk{A}
                    $\batch^r_{k, m} \leftarrow \batch^r_{k, m-1} - \eta_k \cdot \nabla_x(\keloss(\batch^r_{k, m-1}, \theta_k));$
                    \hfill
                    \tikzmk{B}
                    \boxit{black!80}
                }
                each client shares pseudo-gradient $\nabla \batch^{r}_k = \batch^r_{k, M} - \batch^{r}$ with the server;
            }
            // Collaborative knowledge aggregation stage (Eq~\ref{eq:fedream}) \\
            \tikzmk{A}
            $\batch^{r+1}_S \leftarrow \batch^r + \eta_g \sum_{k \in K}\frac{1}{\left|\data_k\right|} \nabla \batch^r_k$;
            \hfill
            \tikzmk{B}
            \\
            \boxit{black!80}
            // Server aggregates model predictions to get $\hat{\data}$ \\
            $\hat{\data} := \{\batch^{r+1}, \hat y^{r+1}_S:= \sum_k\frac{1}{|\data_k|}f_{\theta_k}(\{\batch^{r+1})\}$; \\
            // Local knowledge acquisition stage (Eq \ref{eq:k-acq})\\
            \For{\text{each client} $k \in K$ \textbf{in parallel}}{
                LocalUpdate($\hat{\data}, \theta_k$); LocalUpdate($\data_k, \theta_k$); 
            }
            LocalUpdate($\hat{\data}, \theta_s$);
            
        }
    }
    \caption{\codream Algorithm}
    \label{alg:codream}
\end{algorithm}

\definecolor{LightRed}{rgb}{0.840, 0.529, 0.529}
\definecolor{LightGreen}{rgb}{.632, 0.800, 0.649}

\begin{table*}[]
\resizebox{\textwidth}{!}{
\begin{tabular}{c|ccc|ccc|ccc}

 & \multicolumn{3}{c|}{\textbf{MNIST}} & \multicolumn{3}{c|}{\textbf{SVHN}} & \multicolumn{3}{c}{\textbf{CIFAR10}} \\ 
 
Method & iid($\alpha=\inf$) & $\alpha=1$ & $\alpha=0.1$ & iid($\alpha=\inf$) & $\alpha=1$ & $\alpha=0.1$ & iid($\alpha=\inf$) & $\alpha=1$ & $\alpha=0.1$ \\ \hline \hline
\rowcolor{LightGreen!30}
Centralized & 85.0${}_{(0.9)}\;\;$ & 61.4${}_{(7.1)}\;\;$ & 36.9${}_{(7.6)}\;\;$  & 80.8${}_{(1.3)}\;\;$  & 75.6${}_{(1.4)}\;\;$  & 54.6${}_{(13.6)}\;\;$  & 65.7${}_{(2.9)}\;\;$  & 65.3${}_{(0.4)}\;\;$  & 45.5${}_{(6.8)}\;\;$ \\ 
\rowcolor{LightRed!30}
Independent & 52.4${}_{(7.0)}\;\;$  & 36.3${}_{(6.2)}\;\;$  & 22.0${}_{(4.2)}\;\;$ & 51.3${}_{(9.2)}\;\;$  & 42.3${}_{(6.4)}\;\;$  & 19.6${}_{(9.2)}\;\;$  & 46.4${}_{(2.0)}\;\;$  & 39.7${}_{(3.4)}\;\;$ & 23.5${}_{(5.2)}\;\;$ \\ \hline
\addlinespace
FedAvg & 84.7${}_{(1.6)}\;\;$  & 60.3${}_{(3.4)}\;\;$  & 40.0${}_{(6.9)}\;\;$  & 82.9${}_{(0.4)}\;\;$  & 79.1${}_{(0.9)}\;\;$  & 47.1${}_{(23.7)}\;\;$  & 67.2${}_{(0.4)}\;\;$  & 62.3${}_{(0.9)}\;\;$  & 34.8${}_{(8.3)}\;\;$  \\ 
FedProx & 78.6${}_{(3.5)}\;\;$ & 62.6${}_{(3.6)}\;\;$ & 38.1${}_{(11.0)}\;\;$ & \textbf{86.9${}_{(0.1)}\;\;$} & 84.3${}_{(0.6)}\;\;$ & \textbf{48.7${}_{(26.7)}\;\;$} & \textbf{70.8${}_{(1.8)}\;\;$} & 62.3${}_{(2.9)}\;\;$ & 27.1${}_{(9.8)}\;\;$ \\ 
Moon & 85.1${}_{(2.6)}\;\;$ & 66.2${}_{(4.4)}\;\;$ & \textbf{42.3${}_{(11.8)}\;\;$} & 80.1${}_{(0.1)}\;\;$ & 76.5${}_{(1.2)}\;\;$ & 41.7${}_{(21.8)}\;\;$ & 66.6${}_{(1.4)}\;\;$ & 64.8${}_{(0.8)}\;\;$ & 35.5${}_{(10.8)}\;\;$\\ 
AvgKD & 61.3${}_{(2.3)}\;\;$ & 44.3${}_{(4.8)}\;\;$ & 21.4${}_{(4.3)}\;\;$ & 75.4${}_{(0.7)}\;\;$ & 61.2${}_{(4.6)}\;\;$ & 20.7${}_{(10.9)}\;\;$ & 54.2${}_{(0.9)}\;\;$ & 46.4${}_{(3.3)}\;\;$ & 25.9${}_{(6.2)}\;\;$ \\ 
SCAFFOLD & \textbf{87.5${}_{(0.6)}\;\;$} & \textbf{70.2${}_{(3.6)}\;\;$} & 38.8${}_{(13.7)}\;\;$ & 86.0${}_{(0.1)}\;\;$ & \textbf{84.5${}_{(0.7)}\;\;$} & 13.5${}_{(4.4)}\;\;$ & 73.9${}_{(1.5)}\;\;$ & \textbf{67.5${}_{(4.6)}\;\;$} & 22.8${}_{(7.8)}\;\;$\\ 
FedGen & 64.5${}_{(1.9)}\;\;$ & 51.0${}_{(4.3)}\;\;$ & 31.4${}_{(7.4)}\;\;$ & 49.7${}_{(1.6)}\;\;$ & 44.2${}_{(4.1)}\;\;$& 34.9${}_{(19.7)}\;\;$ & 66.2${}_{(0.4)}\;\;$ & 62.8${}_{(1.8)}\;\;$ & \textbf{40.2${}_{(9.0)}\;\;$} \\ \hline

\codream (ours) & 80.6${}_{(0.5)}\;\;$ & 57.7${}_{(3.6)}\;\;$ & 35.7${}_{(9.2)}\;\;$ & 81.4${}_{(0.1)}\;\;$ & 80.1${}_{(0.8)}\;\;$ & 44.5${}_{(17.7)}\;\;$ & 69.5${}_{(0.3)}\;\;$ & 64.8${}_{(0.3)}\;\;$ & 36.6${}_{(8.4)}\;\;$\\ \hline \hline
\end{tabular}%
}
\vspace{-5pt}
\caption{Performance overview of different techniques with different data settings. A smaller $\alpha$ indicates higher heterogeneity.}
\vspace{-5pt}
\label{tab:non_iid_benchmark}
\end{table*}

\section{Analysis of \feddream}
\label{sec:method-benefits}
The benefits of \feddream are inherited from using KD, along with additional advantages arising from our specific optimization technique. \feddream extracts the knowledge from clients in \textit{dreams} and shares the updates of these dreams instead of model gradients ($\nabla_\theta$) as done in FL.

\textbf{Communication Analysis: }
We use the following notation: $d$ is the dimension of the inputs or \textit{dreams}, $n$ is the batch size of \textit{dreams} generated, and $R$ is the number of aggregation rounds. Since \feddream communicates input gradients ($\nabla_{\batch}$) instead of model gradients ($\nabla_\theta$), the total communication is $d \times n \ R$. In FedAvg and its variants, the communication is $|\theta| \times R$. Unlike in \fedavg, the communication of \feddream is independent of the size of the model parameters $|\theta|$ and remains constant even if the model increases in depth and width. Thus, the communication complexity of \codream does not scale with larger models. For heavily parameterized models, $d\times n \ll |\theta|$. Table \ref{tab:comm_analysis} provides a comprehensive communication analysis for different model architectures in \fedavg vs \feddream.

\textbf{Privacy Analysis: } Exchange of models between the server and clients can result in potential privacy leakage. Various model inversion and reconstruction attacks \cite{haim2022reconstructing, hitaj2017deep} have been shown to leak private sensitive information by reconstructing the training data. However, in \feddream, the clients collaborate by sharing the gradients of \textit{dreams'} without even sharing their model parameters. A simple application of data processing inequality shows that dreams obtained from a model provably have lower information about raw data than the model. Further, we visually analyze generated \textit{dreams} in Figure \ref{fig:dreams}. While \textit{dreams} enable knowledge-distillation, they do not resemble real data. Similar to \fedavg, the synchronization step between the clients is a linear operation (weighted average) and hence offers an additional layer of privacy by using secure and robust aggregation ~\cite{bonawitz2017practical}.

\textbf{Flexibility of models: }
Since the knowledge aggregation in \feddream is done by sharing the updates of \textit{dreams} in data space, \feddream is model agnostic and allows for collaboration among clients with different model architectures. We empirically observe no performance drop in collaborative learning with clients of different model architectures.


\textbf{Customization in sharing knowledge: }Additionally, sharing knowledge in the data space enables adaptive optimization, such as synthesizing adversarially robust samples or class-conditional samples for personalized learning. For more details, refer to the Appendix. \\

\vspace{-10pt}

\vspace{-10pt}
\section{Experiments}
\label{sec:expts}
We conduct our experiments on 3 real-world datasets, including MNIST \cite{lecun1998gradient}, SVHN \cite{netzer2011reading}, and CIFAR10 \cite{krizhevsky2009learning}. Unless stated otherwise, we used ResNet-18~\cite{he2015deep} for training the client and server models and set the total number of clients $K=4$. To validate the effect of collaboration, we train clients with 50 samples per client for MNIST and 1000 samples per client for CIFAR10 and SVHN datasets. 
We refer the reader to the supplementary material for a detailed experimental setup, our code, and the hyperparameters used.
\definecolor{LightRed}{rgb}{0.840, 0.529, 0.529}
\definecolor{LightGreen}{rgb}{.632, 0.800, 0.649}
\begin{table*}[]
\resizebox{\textwidth}{!}{
\begin{tabular}{c|cccc|>{\columncolor{LightRed!30}}c>{\columncolor{LightGreen!30}}cccc}
 & \multicolumn{4}{c|}{\textbf{Heterogeneous Clients (Independent clients 1-4)}} & \multicolumn{4}{c}{\textbf{Method}} \\ 
Model & WRN-16-1 & VGG-11 & WRN-40-1 & ResNet-34 & Independent & Centralized & AvgKD & \codream (ours) \\ \hline \hline
iid($\alpha=\inf$) &  52.2 & 55.1 & 43.5 & 54.2 & 51.6${}_{(4.5)}\;\;$ & 68.8 & 52.9${}_{(1.4)}\;\;$  & 69.6${}_{(1.0)}\;\;$ \\ \hline
$\alpha=1$ & 41.3 & 38.2 & 37.1 &  50.1 & 41.7${}_{(5.1)}\;\;$ & 64.8 & 42.4${}_{(2.9)}\;\;$  & 60.0${}_{(1.7)}\;\;$ \\ \hline
$\alpha=0.1$ & 29.1 & 22.3 & 33.1 & 21.5 & 27.2${}_{(4.9)}\;\;$ & 43.0 & 30.2${}_{(3.3)}\;\;$  & 40.6${}_{(0.9)}\;\;$ \\ \hline
\end{tabular}
}
\vspace{-10pt}
\caption{\textbf{Performance comparison with heterogeneous client models}: on CIFAR10 dataset. Left: Accuracy for independent heterogeneous clients with different models; Right: Average client model performance comparison of \codream with other baselines}
\vspace{-10pt}
\label{tab:het_models}
\end{table*}

\textbf{Client data partition}: To simulate real-world conditions, we perform experiments on both IID and non-IID data. We use Dirichlet distribution $ Dir(\alpha)$ to generate non-IID data partition among labels for a fixed number of total samples at each client. The parameter $\alpha$ guides the degree of imbalance in the training data distribution. A small $\alpha$ generates skewed data. 
Figure \ref{fig:data-distribution} shows data distribution for different $\alpha$. 

\textbf{Baselines}: For baselines, we compare \feddream against \fedavg, FedProx \cite{li2020federated}, Moon\cite{li2021model}, and Scaffold \cite{karimireddy2020scaffold}. We also include other model-agnostic federated baselines such as FedGen\cite{zhu2021data}, which uses a generator model to generate a proxy for locally sensitive data, and AvgKD \cite{afonin2021towards}, which alternately shares models with other clients to get averaged model soft predictions across two clients. We extend the AvgKD method for an n-client setting. More details can be found in the Appendix \ref{appendix:avgkd}. Lastly, we also include \independent and \centralized training baseline for reference. In the \centralized baseline, all the client data are aggregated in a single place. In the case of \independent, we train models only on the client's local dataset and report average client local accuracy. 

\subsection{Fast dreaming for knowledge extraction}
\label{sec:fast_feddream}
Despite the impressive results of the original DreamInversion \cite{yin2020dreaming}, it is found to be extremely slow with 2000 local iterations for a single batch of image generation. The collaborative nature of the knowledge extraction process in \feddream makes it further slow. To accelerate this process of generating \textit{dreams}, the Fast-datafree \cite{fang2022up} approach learns common features using a meta-generator for initializing \textit{dreams}, instead of initializing with random noise every time. This approach achieves a speedup factor of 10 to 100 while preserving the performance. Thus, to speed up our collaborative process of generating \textit{dreams}, we implement \fastfeddream by integrating the Fast-datafree \cite{fang2022up} approach on top of our algorithm. However, in each aggregation round, the client now shares both the local generator model and the dreams for secure aggregation by the server. Instead of 2000 global aggregation rounds (R) in \feddream, \fastfeddream performs only a single global aggregation round with 5 local rounds. We perform all the subsequent experiments using \fastfeddream. More details on the implementation can be found in the Supplement material.

\subsection{Real-world datasets/comparison with FL}
We evaluate our method under both IID and non-IID settings by varying $\alpha = {0.1, 0.5}$ and report the performances of different methods in Table \ref{tab:non_iid_benchmark}. The results show that our approach \feddream achieves high accuracy(close to centralized) across all datasets and data partitions. Even as $\alpha$ becomes smaller (i.e., data become more imbalanced), \feddream still performs well. Note that \codream does not beat other state-of-the-art non-iid techniques since it is not designed for the non-iid data challenges. It is analogous to FedAvg in the data space, and thus, all non-iid tricks can also be applied to \codream to improve its accuracy further.

\subsection{Flexibility of models: Model-agnostic}
Since \feddream shares updates in the data space instead of the model space, our approach is model agnostic. We evaluate our approach across heterogeneous client models having ResNet-34 \cite{he2016deep}, VGG-11 \cite{simonyan2014very}, and Wide-ResNets \cite{zagoruyko2016wide} (WRN-16-1 and WRN-40-1). Table \ref{tab:het_models} shows the performance of \feddream against \centralized, \independent, and model agnostic FL baselines such as Avg-KD. Note that FedGen is not completely model agnostic as it requires the client models to have a shared feature extractor and thus cannot be applied to our setting. We exclude \fedavg as it doesn't support heterogeneous models. Performing FL under both heterogeneous models and non-IID data distribution is a challenging task. Even under this setting, our approach performs better than the baselines.

\subsection{Varying number of clients}
A key goal of \codream is to aggregate knowledge from many decentralized clients. We evaluate this by varying the number of clients $K=[2,4,8,12,24]$, while keeping the total data samples constant. Thus, as $K$ increases, each client contributes fewer local samples.

\begin{figure}[h]
\centering
    \vspace{-5pt}
    \includegraphics[width=0.4\textwidth, height=0.3\textwidth ]{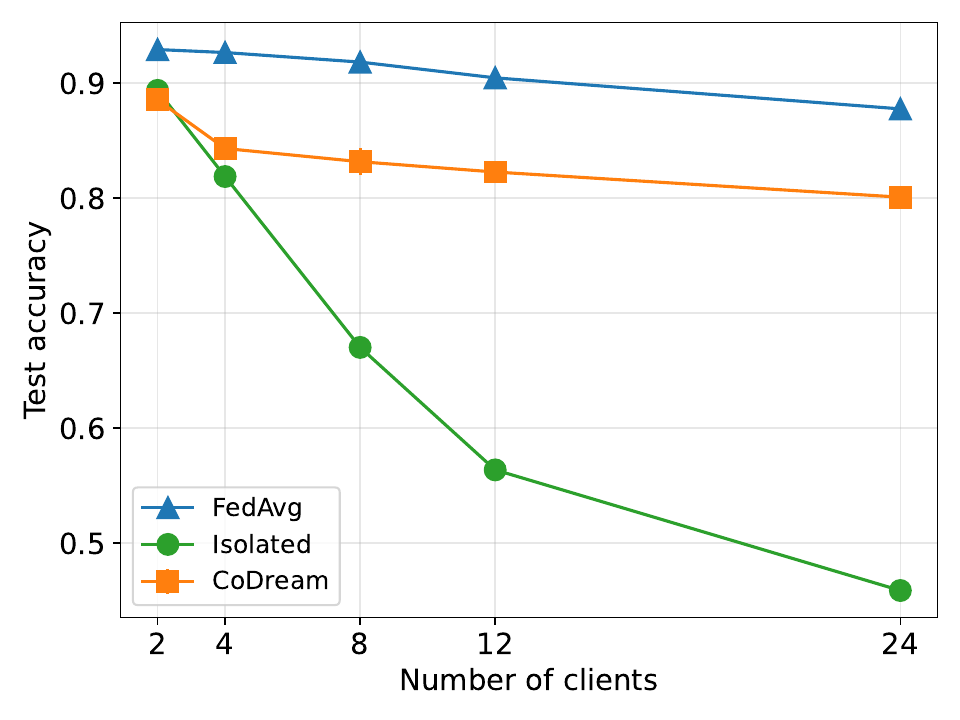}
     \vspace{-10pt}
    \caption{\textbf{Comparison by varying the number of clients.} The performance gap widens between \codream and independent optimization as we increase the number of clients.}
    \label{fig:client_scaling}
\end{figure}

As expected, performance declines with more clients, since each client's knowledge is less representative of the overall distribution. However, Figure \ref{fig:client_scaling} shows this drop is sublinear, making \codream viable for cross-device federated learning. The gap between \codream and \fedavg remains similar across different $K$.

In summary, \codream sees a graceful decline in accuracy as data gets more decentralized. The framework effectively distills collective knowledge, even when local datasets are small. This scalability demonstrates \codream's suitability for privacy-preserving collaborative learning from many heterogeneous client devices.

\vspace{-5pt}

\subsection{Analysis of sample complexity of dreams}
\begin{figure}[h]
    \includegraphics[width=0.45\textwidth, height = 0.3\textwidth ]{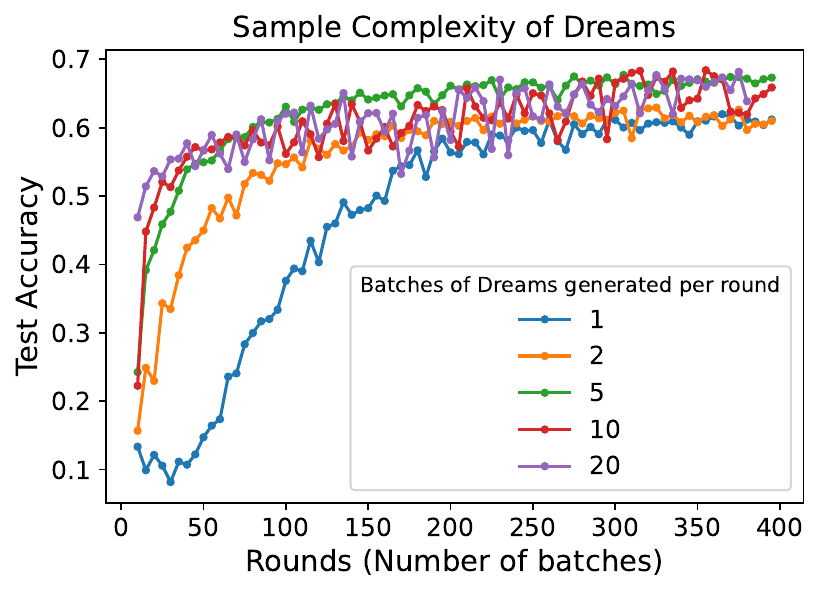}
     \vspace{-15pt}
    \caption{Sample complexity of generated dreams for effective knowledge transfer}
    \vspace{-5pt}
    \label{fig:sample_complexity}
\end{figure}

We plot the accuracy of the server model optimized from scratch against the number of batches of \textit{dreams} it is trained on as shown in Fig~\ref{fig:sample_complexity}. Note that the quality of generated dreams for training increases as training progresses in each round. We also vary the number of batches generated in each round and find 5 batches per round to be an optimum number, after which the marginal gain is very small.
\vspace{-10pt}
\subsection{Validating knowledge-extraction based on  Eq~\ref{eq:knowledge_extraction}}
\begin{figure}[h]
    \centering \includegraphics[width=0.45\textwidth]{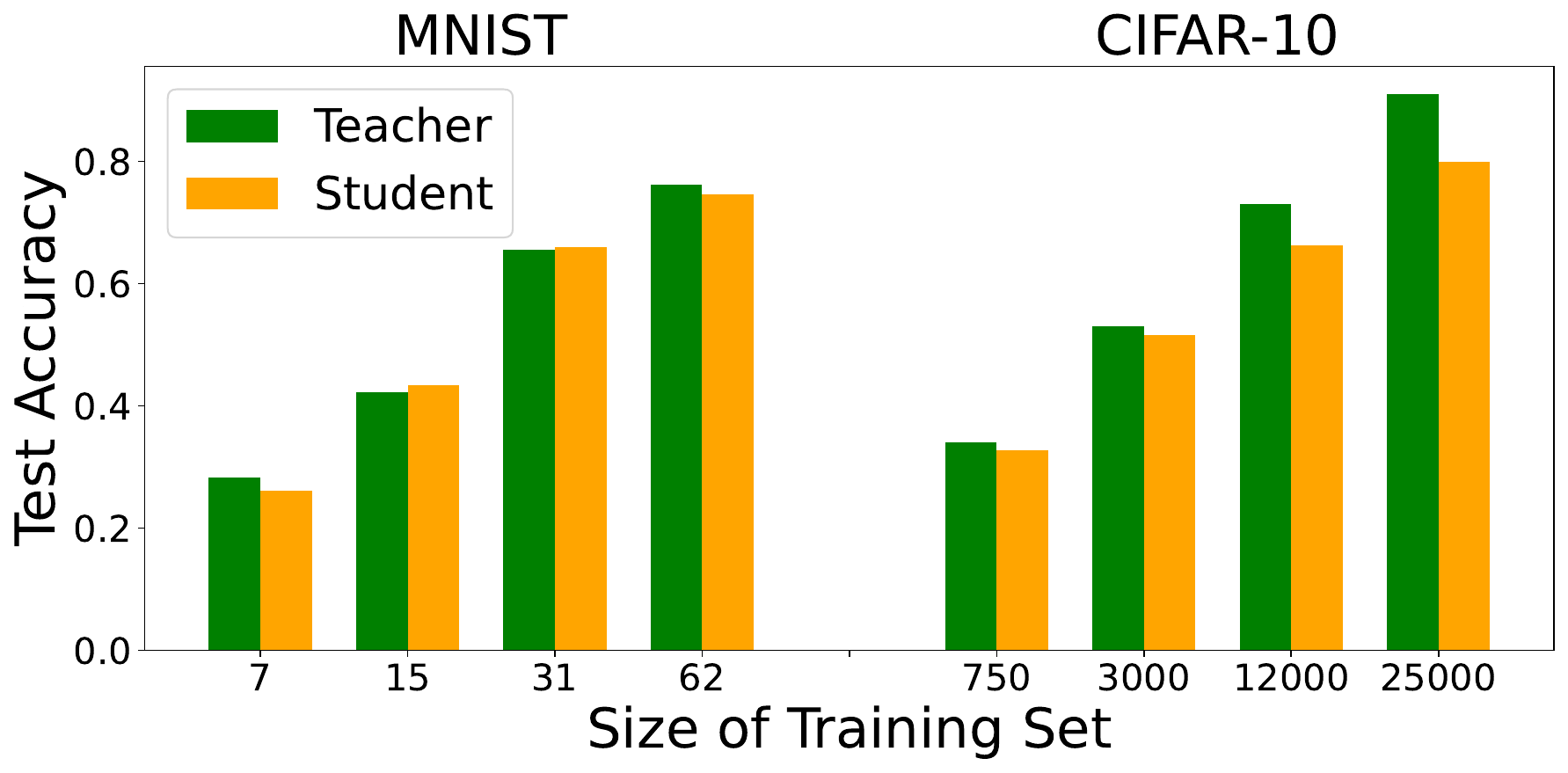}
    \vspace{-10pt}
    \caption{\textbf{Validating the effectiveness of knowledge transfer from teacher to student}: We vary the size of the training dataset (on the x-axis) for the teacher and compare its accuracy with the student trained on dreams generated using Eq~\ref{eq:knowledge_extraction}}
    \label{fig:data_scaling}
\end{figure}

We evaluate whether the knowledge-extraction approach (Sec~\ref{sec:method:k-ex}) allows for the effective transfer of knowledge from teacher to student. We first train a teacher model from scratch on different datasets, synthesize samples with our knowledge-extraction approach, and then train a student on the extracted \textit{dreams}. To validate its compatibility within an FL setting where clients have a small local dataset, we reduce the size of the training set of the teacher to reduce its local accuracy and evaluate how this affects student performance. Results in Fig~\ref{fig:data_scaling} show that the teacher-student performance gap does not degrade consistently even when the teacher's accuracy is low. This result is interesting because the extracted features get worse in quality as we decrease the teacher accuracy, but the performance gap is unaffected.
\vspace{-15pt}

\subsection{Validating collaborative optimization based on Eq \ref{eq:fedream}}
\label{sec:ind_vs_collab}
We also evaluate the effectiveness of collaborative optimization of \textit{dreams} over multiple clients in aggregating the knowledge. To do this, we compare the performance of collaboratively optimized \textit{dreams} in \feddream (using Eq~\ref{eq:knowledge_extraction}) with independently optimized \textit{dreams}. As we can see in table \ref{tab:ablation_impact} (last row), the aggregation step in Eq ~\ref{eq:knowledge_extraction} not only helps in secure averaging, leading to more privacy but also improves the performance.
\begin{table}[!h]
    \centering
    \resizebox{\columnwidth}{!}{
    \begin{tabular}{c|ccc}
    \hline
    Data partition & iid & $\alpha=1$ & $\alpha=0.1$ \\ \hline
    \codream  & 69.2${}_{(0.1)}\;\;$ &  61.6${}_{(0.5)}\;\;$ & 45.6${}_{(1.5)}\;\;$ \\ \hline
    w/o $\mathcal{R}_{adv}$ & 65.7${}_{(0.2)}\;\;$ & 58.4${}_{(1.3)}\;\;$ & 42.0${}_{(1.4)}\;\;$\\ \hline
    w/o $\mathcal{R}_{bn}$ & 51.2${}_{(6.1)}\;\;$ & 33.1${}_{(7.1)}\;\;$ & 24.1${}_{(5.2)}\;\;$\\ \hline
    w/o collab & 64.4${}_{(0.5)}\;\;$ & 58.4${}_{(1.4)}\;\;$ & 30.8${}_{(3.2)}\;\;$\\
    \hline
    \end{tabular}
    }
     \vspace{-10pt}
    \caption{Ablation of components in \feddream on CIFAR10}
    \label{tab:ablation_impact}
    \vspace{-15pt}
\end{table}

\subsection{Contribution of loss components $\mathcal{R}_{bn}$ and $\mathcal{R}_{adv}$ in knowledge extraction}
We further explore the impacts of various components of loss function in data generation in Eq. \ref{eq:knowledge_extraction}. Through leave-one-out testing, we present results by excluding $\mathcal{R}_{bn}$ (w/o $\mathcal{R}_{bn}$) and excluding $\mathcal{R}_{adv}$ (w/o $\mathcal{R}_{adv}$). Table \ref{tab:ablation_impact} shows removing either component influences the accuracy of the overall model, illustrating the impact of each part of the loss function plays an important role in generating good quality \textit{dreams}. We further explain the role of adaptive teaching in Supplement section \ref{sec:adaptive_teaching}.

\vspace{-5pt}
\subsection{Communication efficiency} We compare the client communication cost of \codream and \fedavg per round for different model architectures in Table \ref{tab:comm_analysis}.  In \fedavg, the clients share the model with the server, whereas, in \feddream, they share the \textit{dreams}(size of data).
However, in \codream, each batch of \textit{dreams} is refined for 400 rounds, whereas in \fastfeddream there is only a single round of aggregation along with the sharing of a lightweight generator model (as explained in Section \ref{sec:fast_feddream}.
The communication of both \feddream and \fastfeddream is model agnostic and does not scale with large models. 
We do a further analysis of the tradeoffs between global communication and accuracy of the \feddream approach in Supplementary section \ref{tab:fedopt}.
\begin{table}[h]
    \centering
    \resizebox{\columnwidth}{!}{
    \begin{tabular}{c|ccc}
    \hline
    Model & FedAvg & \feddream & \fastfeddream \\ \hline
    Resent34 & 166.6 MB & 600 MB & 23.5MB \\ \hline
    Resnet18 & 89.4 MB & 600 MB & 23.5MB \\ \hline
    VGG-11 & 1013.6 MB  & 600 MB & 23.5MB \\ \hline
    WRN-16-1 & 1.4 MB & 600 MB & 23.5MB \\ \hline
    WRN-40-1 & 4.5 MB & 600 MB & 23.5MB \\ \hline
    \end{tabular}
    }
     \vspace{-10pt}
    \caption{Communication analysis of \fedavg vs \feddream and \fastfeddream per round}
    \label{tab:comm_analysis}
    \vspace{-10pt}
\end{table}

\subsection{Visual representation of dreams}
\begin{figure}[h]
    \vspace{-5pt}
    \centering
    \includegraphics[width=0.45\textwidth]{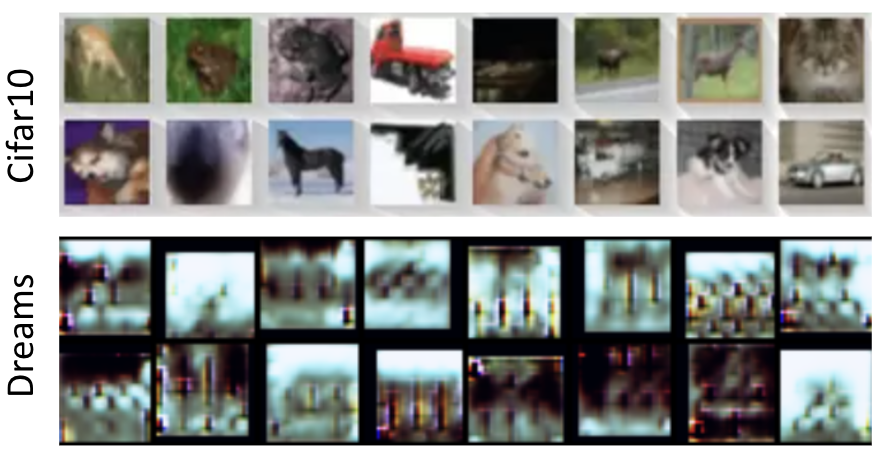}
     \vspace{-10pt}
    \caption{Visualization of \textit{dreams} generated on CIFAR10 dataset}
    \label{fig:dreams}
\end{figure}

Figure \ref{fig:dreams} visualizes the \textit{dreams} generated by \fastfeddream on CIFAR10. While not visually similar to the original training data, these \textit{dreams} effectively encapsulate collaborative knowledge. The goal is to enable decentralized knowledge transfer, not reconstruct the raw data. Thus, models trained on \textit{dreams} perform well despite their visual differences from the underlying distribution. 

\vspace{-20pt}
\section{Limitations and Future work} 
Despite its promising potential, \feddream has some limitations, especially additional computation on client devices. While the number of parameters on the client device remains unchanged, the client device has an additional computation burden. To circumvent this challenge, we implement \fastfeddream which uses a meta-generator that learns good initialization for \textit{dreams} instead of random initialization. Further research and optimizations may be needed to address this limitation. Another promising future avenue is new privacy mechanisms catered for \codream that improve the privacy-utility trade-off.

\section{Conclusion}\vspace{-0.1cm}
In this paper, we introduce \codream, a collaborative data synthesis approach where clients jointly optimize synthetic \textit{dreams} in a privacy-preserving manner. It is a model-agnostic learning framework that leverages a knowledge extraction algorithm by performing gradient descent in the input space. We view this approach as a complementary technique to \fedavg, which performs gradient descent over model parameters. Through comprehensive evaluations and ablation studies, we validate the effectiveness of our proposed method.

\section{Overall Impact}

The proposed \feddream framework significantly advances the landscape of federated learning by introducing key technical innovations with far-reaching implications. Its model-agnostic approach allows clients with diverse architectures to collaboratively optimize data representations, overcoming the need for consensus on model structure. This not only broadens the scope of federated learning but also caters to resource-constrained clients, potentially fostering increased participation in federated ecosystems. The scalability of \feddream, with communication independent of model parameters, addresses concerns related to the size of machine learning models, opening avenues for the deployment of federated learning in scenarios involving large models.

\feddream holds potential across sectors such as healthcare and finance, where data is often decentralized among different entities. By facilitating collaborative data synthesis without centralizing raw information, \feddream supports the development of robust and accurate machine learning models. \feddream's privacy-preserving features, including the two-fold privacy protection and compatibility with secure aggregation, ensure responsible and privacy-aware practices in the context of federated learning. Moreover, by enabling the synthesis of data without direct data sharing, \feddream addresses the concerns related to data ownership and privacy infringement, which are increasingly critical in the era of data-driven technologies. However, \codream does not fix several issues inherent to collaborative learning such as client dropout, stragglers, formal privacy guarantees, bias, fairness, etc. We believe further research is warranted to explore the effectiveness of \codream under those constraints.

\bibliography{main}
\bibliographystyle{icml2024}

\newpage
\appendix
\onecolumn
\section{A detailed analysis of Related Work}
\begin{figure}
    \centering
    \includegraphics[width=\textwidth]{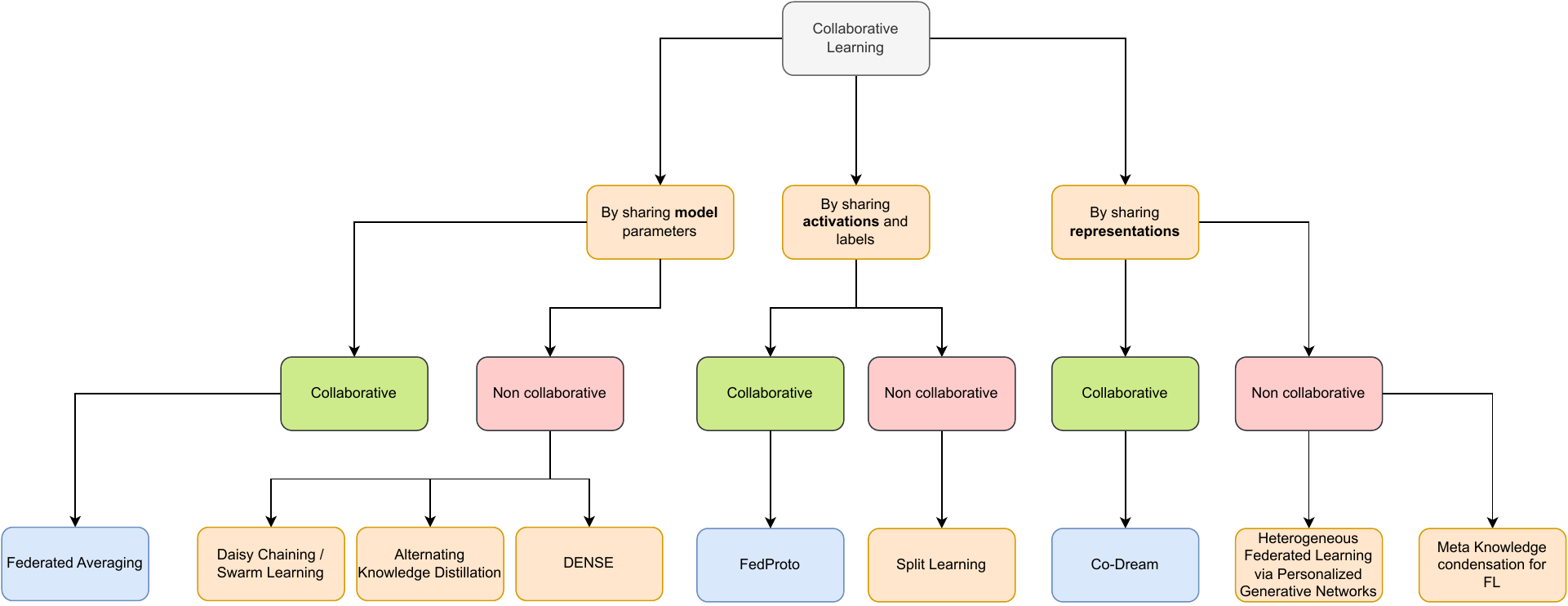}
    \caption{\textbf{Different approaches to collaborative learning}. By collaborative, we mean techniques that perform aggregation of \textit{knowledge} before sharing it with the server.}
    \label{fig:collab_mindmap}
\end{figure}
We illustrate the three classical approaches to collaborative learning in Fig.~\ref{fig:collab_mindmap}. \codream introduces the idea of collaboration in the representation space(data space precisely). From the perspective of our paper, existing works can be categorized based on two characteristics: how knowledge is extracted from the teachers and how the students aggregate and acquire the teachers' knowledge.

\textbf{Knowledge extraction}: One of the key components of KD is to match the output distribution between the teacher and the student conditioned on the same input. Therefore, the same input is needed between the teacher and the student. However, FL techniques typically assume that the student and the teacher have different data siloes. This constraint has led to two general approaches to KD-based FL:
\begin{itemize}
    \item \textbf{Proxy data-based} techniques assume publicly available datasets to obtain the output distribution among different users. Cronus~\cite{chang2019cronus}, Federated model fusion~\cite{lin2020ensemble}, and FedBE~\cite{chen2021fedbe}. Instead of relying on a publicly available dataset, \cite{afonin2022towards} utilize the student's dataset to obtain the teacher's predictions by exchanging the teacher's model. A general drawback of these approaches is that the proxy dataset has to be a sufficiently rich representation of the original data distribution to enable knowledge sharing, which can be an overly constraining assumption for FL.
    \item \textbf{Data Free} techniques learn a generative model of data. DENSE~\cite{zhang2022dense}, Fedgen~\cite{zhu2021data}, and FedFTG~\cite{zhang2022fine} learn a generative model of data on the server. While not focused on knowledge distillation, dataset distillation techniques such as Fed-D3~\cite{song2022federated} and DOS-FL~\cite{zhou2020distilled} perform dataset distillation locally and share the distilled datasets with the server, where a global model is trained with one-shot communication. \cite{jeong2018communication} and \cite{berdoz2022scalable} aggregate the last layer's output on the private data across different samples for every class. A key distinction of our approach from DENSE is that every user shares its model with the server while we assume data synthesis with decentralized models.
\end{itemize}
\textbf{Knowledge Acquisition}: The other key component of KD is the alignment between the teacher's and the student's output distribution $\kl(p(f_{\theta_t}(x))||p(f_{\theta_s}(x)))$ is computed and minimized by optimizing the parameters of the student. Existing approaches can be broadly categorized into where this distribution alignment is performed.
\begin{itemize}
    \item \textbf{Client model regularization} - The clients regularize their local model training by treating other clients as teachers. \cite{lee2021preservation} regularizes on all labels with respect to the aggregated teacher model except the original label in order to avoid catastrophic forgetting. \cite{afonin2022towards} perform distillation locally by using the teacher model on the client's data. \cite{zhu2021data} distribute the synthetic data generator learned by the server to the clients for regularization during training.
    \item \textbf{Ensemble distillation} techniques regularize the global server model by utilizing proxy or synthetic data. FedBE~\cite{chen2021fedbe}, Ensemble Distillation~\cite{lin2020ensemble}, FedAux~\cite{sattler2021fedaux}, and FedFTG~\cite{zhang2022fine} aggregate soft labels from different clients.
\end{itemize}

Independently of knowledge distillation, a few recent works have accommodated model heterogeneity by sub-model extraction~\cite{caldas2018expanding, diao2021heterofl, th2021fjord,alam2022fedrolex} and factorizing model weights with low-rank approximations~\cite{mei2022resource}. Most of these techniques only support heterogeneity for a specific class of models, for example - the ResNet family. In contrast, KD-based approaches are more flexible and only require the input and output dimensionality to be the same.

\section{Complete DeepInversion loss with adaptive teaching}
\label{sec:loss_components}
\subsubsection{Batch-norm regularisation}
To improve the \textit{dreams} image quality, we enforce feature similarities at all levels by minimizing the distance between the feature map statistics for \textit{dreams} and training distribution, which is stored in the batch normalization layers. Hence 
\begin{equation}
    \mathcal{R}_{bn}(\batch) = \sum_l{||\mu_{feat}^l-\mu_{bn}^l|| + ||\sigma_{feat}^l-\sigma_{bn}^l||}
\end{equation}

\subsubsection{Adaptive teaching}
\label{sec:adaptive_teaching}
In passive knowledge transfer, the student does not influence how the teacher extracts knowledge. However, by personalizing the teaching to what a student does not know and what the teacher does know, we can improve the student's performance and speed up learning. Due to gradient-based optimization, the teacher can synthesize examples that maximize the student's loss. The student, in turn, can optimize its model weights to minimize its loss.  Additionally, by personalizing to the student, the teacher avoids extracting redundant knowledge. In our framework, we introduce adaptive teaching at two levels.

Intuitively, for a given teacher, maximizing the differences between the students' and teachers' generation helps generate representations of what the teacher knows and what the student does not. We apply this adaptive teaching technique in \feddream. Under this setup, the clients operate as adaptive teachers for the server and minimize their loss while maximizing the loss with respect to the server. The server's knowledge can be viewed as the culmination of all the clients' knowledge compressed into a single model. The idea of flipping the gradients for learning what you don't know is interesting.

Thus to increase the diversity in generated \textit{dreams}, we add an adversarial loss to encourage the synthesized images to cause student-teacher disagreement. $\mathcal{R}_{adv}$ penalizes similarities in image generation based on the Jensen-Shannon divergence between the teacher and student distribution,
\begin{equation}
    \mathcal{R}_{adv}(\batch) =- JSD(f_t(\batch) || f_s(\batch))
\end{equation}
where the client model is the teacher and the server model is the student model. To do this adaptive teaching in a federated setting, the server shares the gradient $\nabla_{\batch}f_s(\batch)$ with the clients for local adaptive extraction. The clients then locally calculate $\nabla_{\batch}\tilde{\ell}(\batch,\; \theta_k)$ which is then aggregated at the server for knowledge aggregation in Eq \ref{eq:fedream}.

\section{Experimental Setup}
We used PyTorch~\cite{NEURIPS2019_9015} framework for training and evaluation in all of the experiments. All of our experiments were performed over the Nvidia-GeForce GTX TITAN GPU. We start with a few warmup rounds (for example, 20) for each client to update their local model.  We use a batch size of 256 for generating dreams as well as training on local data. We generate 1 batch of \textit{dreams} per round which can be further increased up to 5 after which the marginal increase in accuracy is low. We also maintain a buffer for \textit{dreams} dataloader with a fixed size in which new dreams are added in each round as the local models are updated and the old ones are removed. For optimizing the dreams in \feddream, we used Adam with a learning rate of 0.05. For each round in collaborative learning, there were 2000 global aggregation rounds and 1 local round for optimizing dreams. We do further communication vs accuracy analysis of our \feddream approach in Section \ref{sec:fed_opt}. 
In \fastfeddream, clients collaboratively train a generator model to learn a good initialization for the meta-features. However, we perform 5 local rounds with only one global aggregation round. 
We empirically observe that the optimizers' choice for the dreams and the networks was crucial to obtaining good representations. 
We perform 2 types of training to update the client's local models using SGD with a learning rate of 0.2 and momentum of 0.9. (1) The client models are trained on the local data using cross-entropy loss. (2) To update the models on the global knowledge, each client trains their models on the global dream dataset using knowledge distillation loss. 

\begin{figure*}[h]
    \includegraphics[width=0.32\textwidth,height=0.25\textwidth]{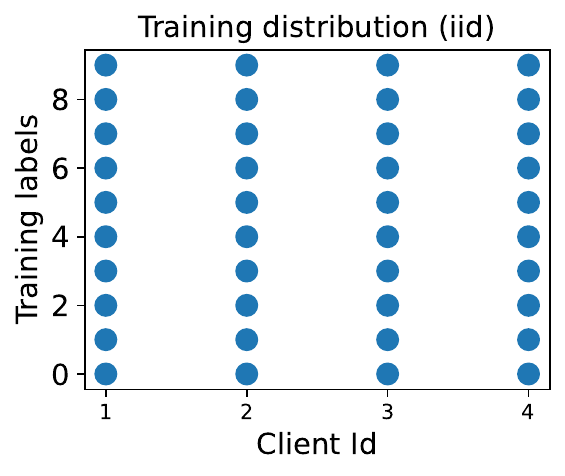}
    \includegraphics[width=0.32\textwidth,height=0.25\textwidth]{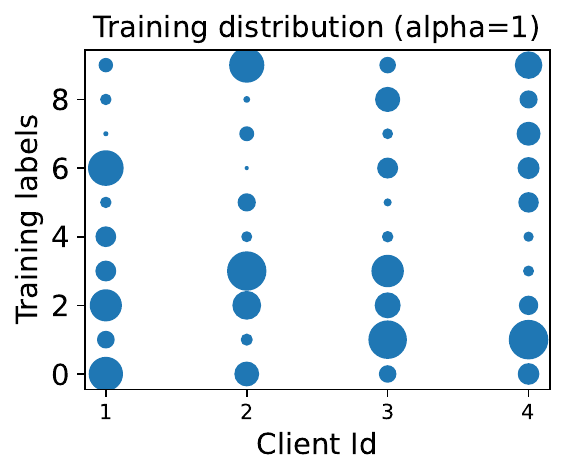}
    \includegraphics[width=0.32\textwidth,height=0.25\textwidth]{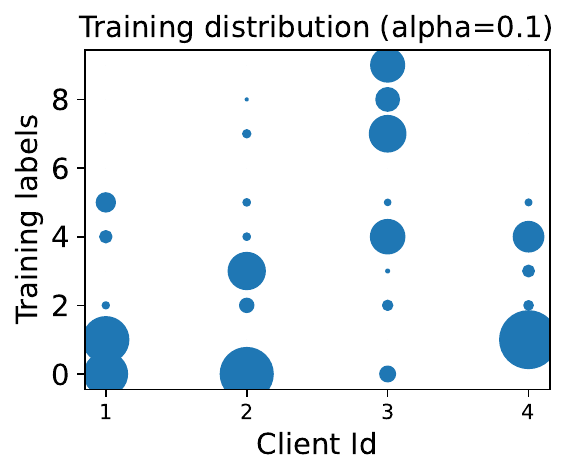}
    \vspace{-15pt}
    \caption{Visualization of statistical heterogeneity among CIFAR10 clients: Client IDs on the x-axis, class labels on the y-axis, and point size reflects the number of training samples per label for each user}
    \label{fig:data-distribution}
\end{figure*} 

Further, in some non-iid data partitions among clients, we observe that updating the client models on the local dataset which might be highly imbalanced diverges the local model from the global optimum. To address this issue, we use a naive approach of turning off the training on the local dataset after certain epochs for the clients who start deviating a lot from the model updates on training on the global dream dataset.

Unless stated otherwise, we used ResNet-18~\cite{he2015deep} for training the client models and the model at the server (for applicable experiments) and a batch size of \textit{256} for generating dreams and for training these models. We have shared our code here \url{https://drive.google.com/drive/folders/1m9UbOQCTqr5Hjtkl4FR9x-peDiZJXg3U?usp=share_link}. Specific hyper-parameters for dream optimization for both approaches can be found in the config file in our code.
\section{Experimental Analysis}
\subsection{Comparison between collaborative and independent optimization}
\begin{figure}[!hbt]
    \centering
    \includegraphics[width=0.9\textwidth]{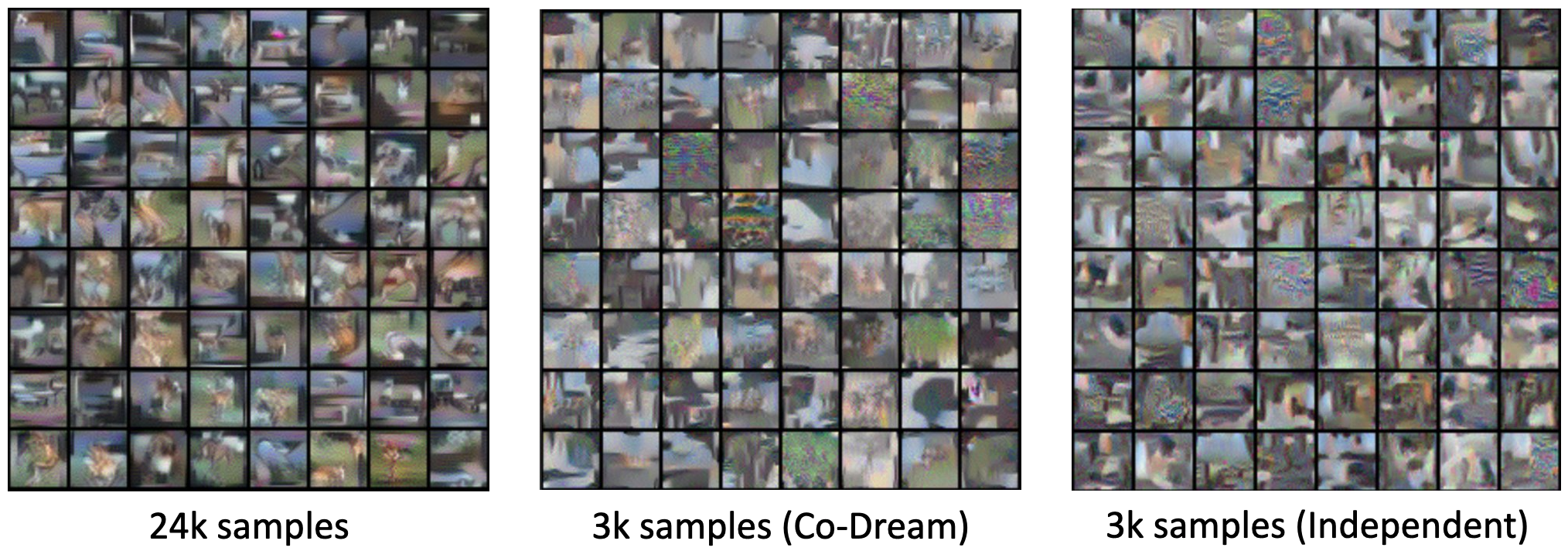}
    \caption{We qualitatively compare the dreams optimized collaboratively using \codream and optimized independently using our knowledge extraction procedure.}
    \label{fig:codreamvsind}
\end{figure}

\subsection{Federated averaging versus distributed optimization in \feddream}
\label{sec:fed_opt}
Similar to \fedavg, our approach reduces client communication by increasing the number of local steps performed on the client device. Therefore, we quantify the tradeoff between the number of local steps and the reduction in the co-dream quality. We note that our knowledge extraction approach is sensitive to the optimizer and usually performs better with Adam~\cite{kingma2014adam} over SGD.
\begin{table}[!hbt]
    \caption{Comparison of different optimization techniques for \codream}
    \centering
    \begin{tabular}{ccccc}
        \toprule
    Optimization & Global rounds (R)  & Local rounds (m) & MNIST & CIFAR10 \\
    \midrule
    DistAdam & 2000 & 1 & 0.763 & 0.644\\
    \midrule
    FedAvg & 400 & 5 & 0.1826 & 0.5919\\
    \midrule
    DistAdam & 400& 5 & 0.7978 & 0.5949\\
    \midrule
    FedAdam & 400 & 5& 0.7831 & 0.6439 \\
    \bottomrule
    \end{tabular}
    \label{tab:fedopt}
\end{table}

This presents a unique challenge when performing multiple local optimization steps locally, as the server can no longer perform adaptive optimization. Therefore, we utilize the same approach as adaptive federated optimization~\cite{reddi2020adaptive} that treats the server aggregation step as an optimization problem and replaces the simple averaging (i.e. FedAvg) with adaptive averaging with learnable parameters on the server.

We compare three methods of optimization: 1)~\textit{DistAdam} where the clients share gradients at every step and the server applies Adam optimizer on the aggregated gradients, 2)~\textit{FedAvg} where clients apply Adam optimizer locally for $m$ steps and the server averages the \textit{pseudo-gradients} as described in Eq~\ref{eq:fedream}, and 3) \textit{FedAdam} where clients apply Adam optimizer locally for $m$ steps and the server performs adaptive optimization on the aggregated \textit{pseudo-gradients} based on the formulation by~\cite{reddi2020adaptive}.

\begin{figure}[H]
    \centering
    \includegraphics[width=0.6\textwidth]{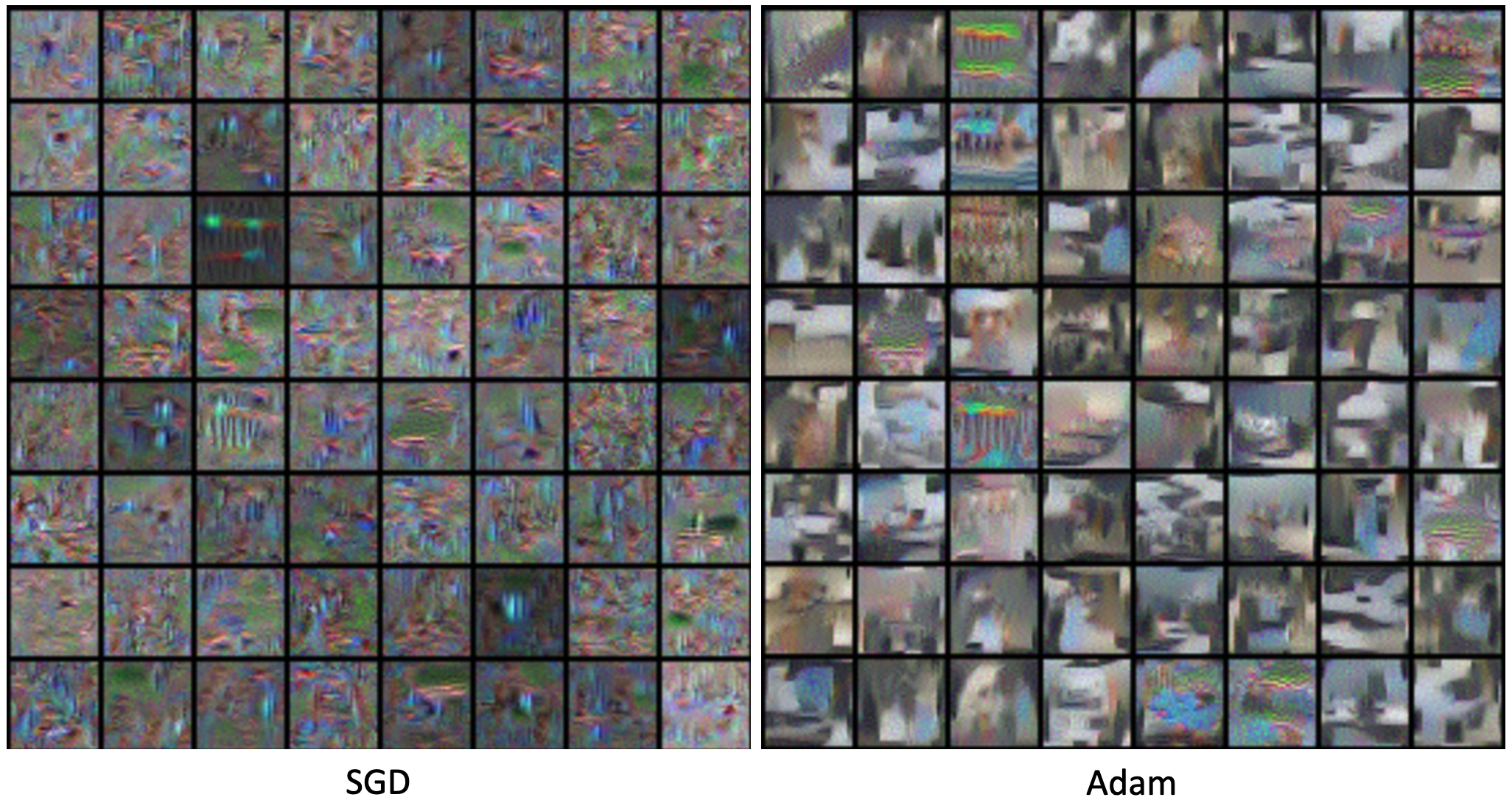}
    \caption{We qualitatively compare the dreams obtained using SGD and Adam optimizers. A key implication of this result is that during \codream the server needs to perform adaptive averaging instead of typical weighted averaging.}
    \label{fig:opt_comp}
\end{figure}

We show qualitative results in the Figure \ref{fig:opt_comp} and quantitative difference in Table~\ref{tab:fedopt}. We find that the naive \textit{FedAvg} approach reduces the student performance even with a minor increase in the number of local computation steps; however, when we apply \textit{FedAdam}~\cite{reddi2020adaptive}, we see similar performance as \textit{DistAdam} with reduced global steps.

\subsection{Consensus across clients}
We look at the predictions by each model on the synthesized dreams. Since each model makes their prediction independently, we hypothesize that similar predictions among clients can be viewed as a consensus among clients. Our results indicate that clients indeed achieve consensus over collaboratively synthesized dreams.
\begin{figure}[!hbt]
    \centering
    \includegraphics[width=\textwidth]{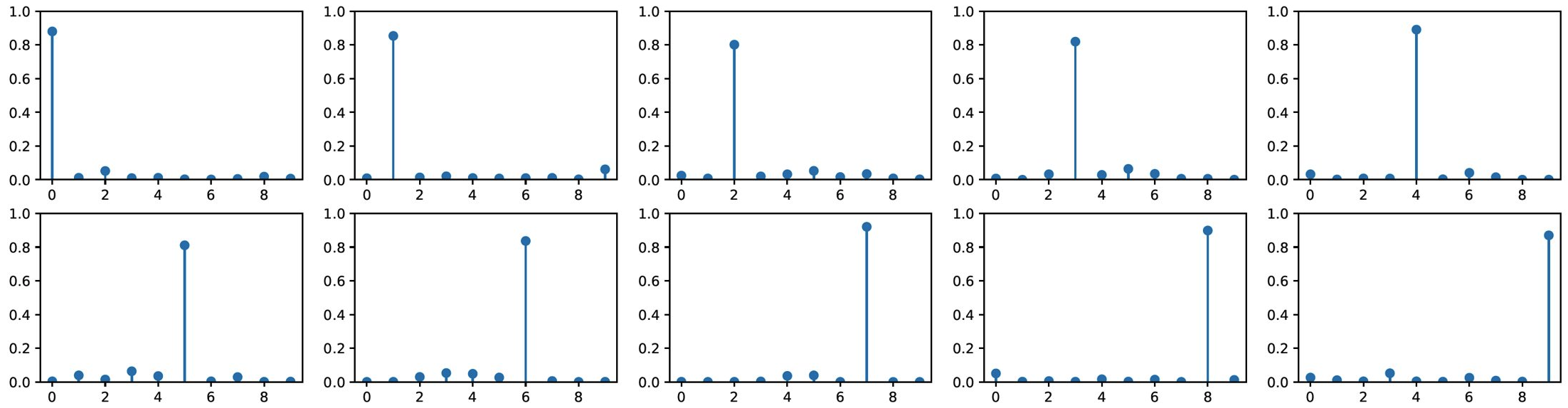}
    \caption{We take dreams obtained using \codream for clients trained on CIFAR-10 and compute the output for each client on every given dream and average it across clients. We then segregate these dreams into 10 buckets based on each dream's corresponding $\argmax$. We plot the output probabilities for each bucket separately. We observe that for each class the clients have sharp peaks. The existence of sharp peaks imply consensus across clients for the dreams because if the clients disagreed on the dreams, then the output classes would be different and hence the distribution would skew more towards a uniform one.}
    \label{fig:my_label}
\end{figure}

\section{Average Knowledge Distillation (AvgKD) Baseline}
\label{appendix:avgkd}
We follow the procedure in \cite{afonin2021towards} while implementing AvgKD for N-client setting. In the first round t=0, each client model is trained on their local data $(X^{i}, y^{i})$ to obtain the local model $g_{0}^{i}$. In each subsequent round t, for each agent i, the model $g_{t}^{i}$ is shared with all the other agents j = 1, ... , N, where j $\neq$ i. Now, every agent uses the shared models to predict the next set of soft labels as follows: 

\[ y_{t+1}^{i} = \frac{y^{i} + \sum_{j=1,j \neq i}^{N} g_{t}^{j} (X^{i})}{N} \]

where $y^{i}$ represents the ground truth labels for client i. Each agent model is now trained on this new set of soft labels. In addition to this AvgKD algorithm, we added a local training step at each round which gave a better performance on all datasets. In this algorithm, only the model weights are shared across agents, and there is no exchange of data. However, since this method requires all-to-all communication, it is not suitable for a higher number of agents and is not scalable. 

Looking at the results for AvgKD for four clients with the same ResNet-18 model for all clients in Table \ref{tab:non_iid_benchmark}, we can see that AvgKD consistently performs better than the independent client training scenario for $\alpha = \infty$ (iid case) and $\alpha = 1$ for all the three datasets, CIFAR10, SVHN, and MNIST. However, this is not true for the highly heterogeneous data case with $\alpha = 0.1$. This coincides with the findings of \cite{afonin2021towards} that the AvgKD algorithm doesn't perform well when the clients have highly heterogeneous datasets.

Further, for Table \ref{tab:het_models}, we look at how AvgKD performs in a heterogeneous model case on the CIFAR10 dataset. The authors in \cite{afonin2021towards} claim that the AvgKD algorithm works well for the heterogeneous model case with two clients. We find that the performance of the 4-client heterogeneous model setting of AvgKD is very similar to the independent heterogeneous case in terms of accuracy for the CIFAR10 dataset.

\end{document}